\documentclass{article}
\usepackage[T1]{fontenc}
\usepackage{iclr2027_conference,times}
\usepackage{amsmath,amssymb}
\usepackage{graphicx,booktabs,wrapfig}
\usepackage{capt-of,flafter}
\usepackage{xcolor,enumitem,url}
\usepackage{tcolorbox}
\tcbuselibrary{listings}
\usepackage{algorithm,algpseudocode}
\usepackage[pdfversion=1.7]{hyperref}

\definecolor{resultgreen}{RGB}{22,118,75}
\definecolor{resultred}{RGB}{174,57,49}

\iclrfinalcopy

\newcommand{\gain}[1]{{\color{resultgreen}#1}}
\newcommand{\loss}[1]{{\color{resultred}#1}}
\newcommand{\modify}[1]{{\color{black}#1}} 

\newcommand{\sg}{\operatorname{sg}}

\title{CodeLift: Your Test Suite Is Secretly a Dense SWE Curriculum}

\author{Junjielong Xu$^{1,2}$ \thanks{Work was done in Tencent HY.} \quad Huachao Zhu$^1$ \quad Alyssa Wu$^2$ \quad Zhichao Hu$^2$ \quad Pinjia He$^{1 \dagger}$\\
$^1$School of Data Science, The Chinese University of Hong Kong, Shenzhen \quad $^2$Tencent\\
\texttt{\{junjielongxu,huachaozhu\}@link.cuhk.edu.cn,}\\
\texttt{\{alyssawwu,frankiehu\}@tencent.com,} \quad \texttt{hepinjia@cuhk.edu.cn}
}

\begin{document}
\raggedbottom
\maketitle

\lhead{Preprint} 

\begin{abstract}
The \textit{pass-all-test} binary reward has become a standard recipe for training software engineering agents, but its sparsity eliminates difference of partially correct patches.
The test \textit{pass-rate} is a natural alternative.
However, we find that it is not a reliable indicator of patch quality.
A
functionality
being tested can be covered by either one challenging test or one hundred trivial tests, where the majority might be already satisfied \textit{pass-to-pass} tests.
 We introduce \textbf{CodeLift}, a reinforcement learning method that dynamically calibrates pass-rate rewards, turning the test suite into an adaptive curriculum.
At each iteration, CodeLift groups tests that kill the same patches and patches that pass the same tests, then gives more weight to test groups that fewer patch groups pass.
This reduces the impact of majority simple tests and focuses more on what is still challenging.
Within each rollout, CodeLift applies the same scoring rule to repository snapshots, assigning credit to useful actions even when later ones introduce regressions.
Experiments on SWE benchmarks show that CodeLift consistently improves pass@1/2/4/8 over binary and pass-rate baselines.
Analyses show that CodeLift provides dense, stable learning signals and unlocks the inherent curriculum in the test suite.
\end{abstract}%

\section{Introduction}

Large language model (LLM) agents have made notable progress in software engineering (SWE) \citep{jimenez2024swebench,jain2025r2egym}, where reinforcement learning with verifiable rewards (RLVR) served as a key contributor~\citep{zeng2026glm,xu2026deepseek}.
To solve an issue, the SWE agents typically navigate the repository, edit code, and run tests, where the final patches are used as the final answer to be verified.
The standard recipe uses \textit{pass-all-test} as a verifier, providing binary reward based on whether the final patch passes all hidden tests.
While it is
sound, its binary value is too sparse to distinguish the patches that partially resolve the issue from those that break existing
functionality
Thus, on challenging problems where %
all patches in a rollout group for the same issue fail, there is no learning signal from relative advantage, making training inefficient.

The test \textit{pass-rate} is a natural denser alternative.
However, a higher pass-rate does not always mean a better patch.
In competitive programming, recent studies show that
the pass-rate does not achieve a higher performance ceiling than the binary reward
, as it may reinforce shortcuts (e.g., brute-force heuristics)~\citep{li2026exploring}.
In SWE agentic coding, we find it even worse, as the pass rate may eventually lead to collapse (Figure~\ref{fig:validation}).
We identify three
test-weighting imbalances
behind such miscalibration.
(1) \textit{\textbf{Multiplicity imbalance}}:
In practice,
different test cases may check the same functionality and fail together when it is violated.
A patch with a higher pass rate may fix only a few
functionalities
covered by many tests, while breaking many others covered by fewer tests.
\modify{Thus, such test multiplicity can introduce reward biases, as a patch satisfying fewer functionalities might get more rewards.}
(2) \textit{\textbf{Progress-regression imbalance}}:
Most repository tests are regression tests that already pass and protect existing functionality~\citep{rothermel1997safe} (\textit{pass-to-pass}, P2P), while only a few hidden tests initially fail and check the requested progress (\textit{fail-to-pass}, F2P).
Under pass-rate rewards, the penalty for introducing P2P regressions can outweigh the reward for progress on F2P tests, discouraging agents from making changes.
(3) \textit{\textbf{Difficulty imbalance}}:
Pass rate assigns equal weight to tests of different difficulty.
When easy tests dominate the suite, a patch can get a high reward while still failing the challenging ones.
In contrast, the progress on the hard tests can only affect a small proportion of the total reward.

\begin{figure}[!th]
  \centering
  \includegraphics[width=\textwidth]{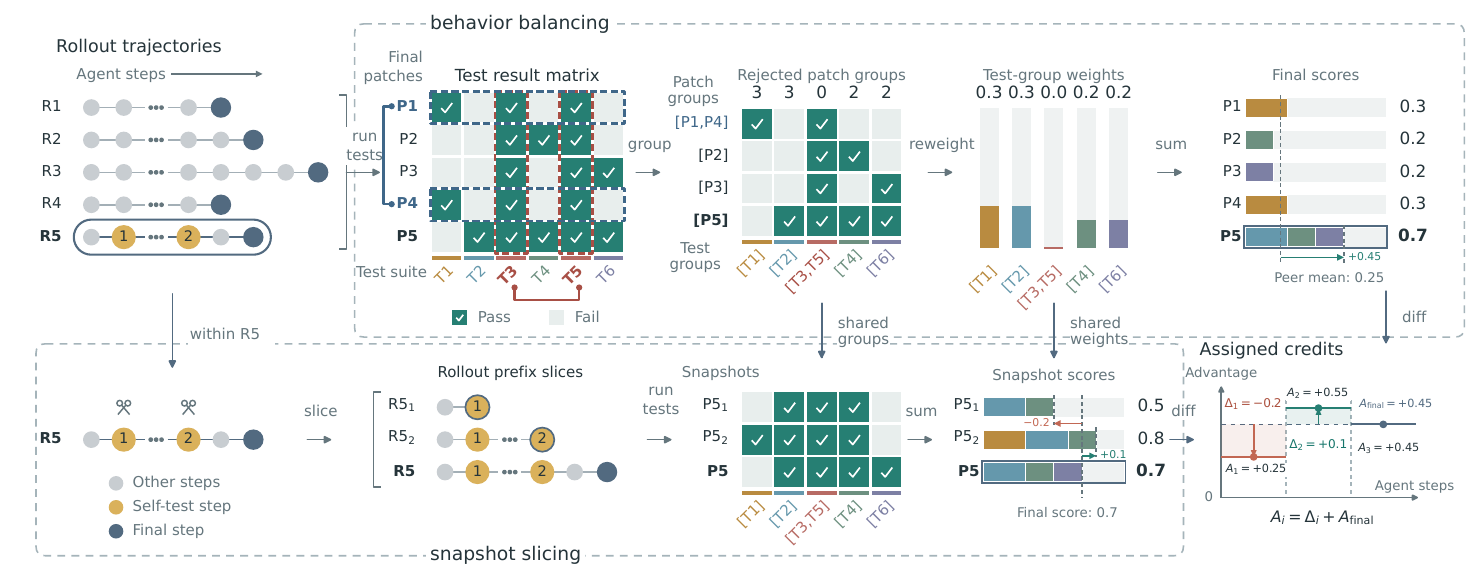}
  \caption[CodeLift overview]{\textbf{Overview of CodeLift.}
[Top] \textit{Behavior balancing}: groups tests and patches by their outcomes, weights test groups by the patch groups they kill, and scores patches using these weights.
[Bottom] \textit{Snapshot slicing}: uses self-tests to select candidate checkpoints, scores intermediate snapshots and combines the snapshot deltas $\Delta_i$ with final advantage $A_{\mathrm{final}}$ to obtain slice advantage $A_i$.}
  \label{fig:overview}
\end{figure}%

We introduce \textbf{CodeLift}, a reinforcement learning method that dynamically calibrates pass-rate through \textit{behavior balancing} at each training step.
The core idea is to use observed behavior of tests and patches to reduce pass-rate's weighting imbalance by grouping tests killing the same patches and patches passing the same tests.
Inspired by \textit{mutation testing theory}~\citep{demillo1978hints}, CodeLift
uses different patch groups as empirical contrast sets, and estimates each test group's current difficulty by the fraction of patch groups they kill.
After that, it normalizes these weights across test groups, and then computes each patch's reward as the sum of the weights of the groups it passes.
This process naturally introduces an \textit{adaptive curriculum}: all-pass tests and all-fail tests are classified to two distinct groups, while tests with distinguishable behaviors are classified into separate groups.
As the policy improves during training, CodeLift updates the groups and their weights, giving less weight to easier groups and more to harder ones.
For example, all-pass tests receive zero weight, with the remaining weights renormalized to keep scores in $[0,1]$ and keep \modify{weight} sum to $1$.
If some patches later regress, the affected tests regain weight, rewarding patches that still pass and lowering the relative advantages of those that fail.

CodeLift further refines credit assignment through \textit{snapshot slicing}, applying the same scoring rule used for final patches to intermediate repository states.
The intuition is: when agents test their edits, they might be checking whether current changes work, making self-testing a natural checkpoint for evaluating the progress so far.
If a rollout ended at that point, its cumulative edits would form a candidate final patch.
CodeLift compares the snapshot's score with the final patch's score, adding the relative difference to the final advantage for the corresponding slice.
Thus, it can assign credit to intermediate actions with progress even when later steps introduce regressions.

We evaluate CodeLift on SWE-bench Verified~\citep{jimenez2024swebench} and SWE-bench Pro~\citep{deng2025swebenchpro} Python subset, using Qwen3-30B-A3B-Instruct-2507 (Qwen3)~\citep{qwen3technicalreport} trained on R2E-Gym~\citep{jain2025r2egym}.
CodeLift consistently outperforms binary and enhanced pass-rate reward baselines across pass@1/2/4/8, \modify{with gains ranging from 2.75 to 7.40 on Verified and from 0.22 to 4.51 on Pro.}
Our analyses show that CodeLift provides denser learning signals, more balanced rewards, and adaptive curriculum driven by the policy's test behavior at each training step.

Our contributions are as follows:
\begin{itemize}[leftmargin=*,label=$\bullet$]
  \item We identify three potential sources that miscalibrate pass-rate rewards for SWE agents: (1)
  multiplicity
  imbalance, (2) progress--regression imbalance, and (3) difficulty imbalances.
  \item We introduce \textbf{CodeLift} \modify{with two key components: (1) \textit{behavior balancing} and (2) \textit{snapshot slicing}, which turns test results into dense rewards with adaptive curriculum and credit assignment}.
  \item We show CodeLift's consistent gains over binary and enhanced pass-rate baselines on SWE benchmarks, and provide detailed analysis for its characteristics.
\end{itemize}

\section{Learning from Repository Tests}
\label{sec:preliminaries}

\paragraph{SWE Agentic RL.}
Given an issue $x$ with $m_x\ge1$ golden tests, an agent samples a rollout $\tau\sim\pi_\theta(\cdot\mid x)$ by interacting with a repository environment through tool-call actions and tool-response observations.
The final repository state $s_{\mathrm{final}}(\tau)$ is captured as a patch.
A common objective is to maximize the probability that the final patch passes all required tests \citep{jain2025r2egym}:
\begin{equation}
  J_{\mathrm{success}}(\theta)
  =\mathbb E_{x,\tau}\!\left[R_{\mathrm{Binary}}(s_{\mathrm{final}}(\tau))\right],
  \qquad R_{\mathrm{Binary}}(s)=\prod_{t=1}^{{m_x}}o_t(s),
  \label{eq:success}
\end{equation}
where $o_t(s)\in\{0,1\}$ is the result of the test $t$. All patches with failed tests get zero reward regardless of partial progress in this binary setting~\citep{li2026exploring}.

\paragraph{Imbalances in Pass-Rate Rewards.}
A natural dense reward is the fraction of passing tests:
\begin{equation}
  R_{\mathrm{PassRate}}(s)=\frac{1}{{m_x}}\sum_{t=1}^{{m_x}}o_t(s).
  \label{eq:pass-rate-equation}
\end{equation}
However, counting tests equally for pass-rate rewards faces three test-weighting imbalances:

\begin{figure}[!htb]
  \centering
  \includegraphics[width=\textwidth]{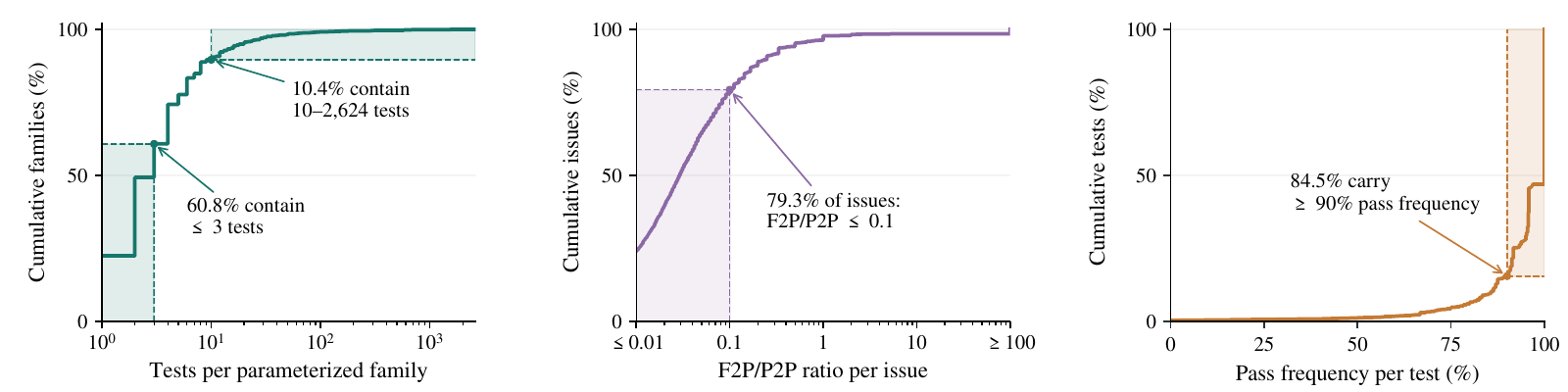}
  \caption[Test-suite distributions]{\textbf{%
Cumulative test-suite distributions on R2E-Gym.}
[Left] Tests per parameterized family.
[Middle] F2P/P2P per issue distribution.
[Right] final-patch test pass frequency distribution.}
  \label{fig:miscalibration}
\end{figure}%

\begin{itemize}[leftmargin=*]
\item \textit{Multiplicity imbalance}:
In practice, diverse test cases can check the same functionalities under different inputs or conditions.
However, it is challenging to determine the functionalities each test covers without human labeling.
Though \textit{parameterized tests} illustrates which tests are explicitly used for the same functionalities, they only cover a subset of test suite (e.g., 43.0\% in R2E-Gym), and other non-parameterized tests may also cover the same functionalities.
Figure~\ref{fig:miscalibration} (left) shows a long-tailed distribution: most
parameterized familes
are covered by very few tests, while few contain far more tests
and higher impact on pass rate.
\modify{Thus, using raw pass-rate could introduce a flip in reward ordering: a patch satisfying fewer heavily tested functionalities might get a higher score than those achieveing more lightly-tested functionalities.}
\item \textit{Progress--regression imbalance}:
For an SWE task, there are two types of tests:
\textit{Pass-to-pass} (P2P) tests check whether existing functionality is broken, which automatically passes in the original repository without changes and should keep passing after patch.
\textit{Fail-to-pass} (F2P) tests check whether requested changes are implemented, which initially fail but should pass if the patch is correctly implemented.
Figure~\ref{fig:miscalibration} (middle) shows that in most issues, P2P tests much more than F2P tests for most issues.
Thus, most of the pass-rate reward comes from keeping existing functionality, leaving only a small share for implementing the requested changes.
\item \textit{Difficulty imbalance}:
It is challenging to find a ground-truth label for test's difficulty.
An empirical method is to estimate it via directly using the pass rate on each test.
Figure~\ref{fig:miscalibration} (right)
shows that most tests pass on nearly all sampled patches.
Thus, a patch can get a high pass-rate reward by passing all trivial tests while still failing on each challenging test.
In contrast, progress on hard tests has a small impact on total reward.
\end{itemize}

\section{CodeLift}
\label{sec:method}

To reduce these test-weighting imbalances above, we propose CodeLift, which turns test outcomes into dense rewards with an adaptive curriculum.
CodeLift contains two major components: behavior balancing and snapshot slicing:
Behavior balancing defines the scoring rule, while snapshot slicing applies it within each rollout to refine credit for intermediate progress. We summarize the whole process of CodeLift in Algorithm~\ref{alg:sbr-training} and illustrate it with an dummy example in Figure~\ref{fig:overview}.

\begin{algorithm}[!htb]
\caption{CodeLift policy update}
\label{alg:sbr-training}
\textbf{Input:} policy $\pi_\theta$; issue batch $\mathcal X$; rollouts per issue
$n\ge2$; each issue $x$ has $m_x\ge1$ golden tests.
\begin{algorithmic}[1]
\State $\theta_{\rm old}\gets\theta$; $\mathcal T\gets[\,]$
\For{each issue $x\in\mathcal X$}
  \State Collect $\tau^1,\ldots,\tau^{n}$ from rollout workers for $x$
  \State Test final patches: ${M_{jt}}\gets o_t(s_{\mathrm{final}}(\tau^j))$, $M\in\{0,1\}^{{n\times m_x}}$ \Comment{\textbf{Behavior balancing}}
  \Statex \hspace{\algorithmicindent}\ \ (where ${M_{j\cdot}}$ are patch vectors across tests; ${M_{\cdot t}}$ are test vectors across patches)
  \State $\mathcal G_{\mathrm{patch}}\gets{\operatorname{GroupEqualRows}}(M)$; $\mathcal G_{\mathrm{test}}\gets{\operatorname{GroupEqualColumns}}(M)$
  \State Compute test-group weights $w_g$ by Eq.~\ref{eq:weights}
  ($w_g=0$ when all patch pass test group $g$)
  \State Compute all $r_{\mathrm{final}}^j=S(s_{\mathrm{final}}(\tau^j))$, $b_j$, and $A_{\mathrm{final}}^j$ by Eq.~\ref{eq:slice-rloo}
  \For{each rollout $j=1,\ldots,{n}$}
    \State $\widehat A_k^j\gets A_{\mathrm{final}}^j$ for every generated token $k=1,\ldots,L_j$
    \If{rollout $\tau^j$ contains self-tests} \Comment{\textbf{Snapshot slicing}}
      \State $(s_i^j,I_i^j)_{i=1}^{K_j}\gets\operatorname{SelfTestSlice}(\tau^j)$ (Sec.~\ref{sec:slicing})
      \State Compute $S(s_i^j),\Delta_i^j,A_i^j$ by Eqs.~\ref{eq:score}--\ref{eq:general-advantage} with fixed $w_g$
      \State $\widehat A_k^j\gets A_i^j$ for $k\in I_i^j$
    \EndIf
    \State Append $(\tau^j,(\widehat A_k^j)_{k=1}^{L_j})$ to $\mathcal T$
  \EndFor
\EndFor
\State If $\mathcal T\ne[\,]$, optimize $\theta$ on $\mathcal T$ with fixed credits and $\theta_{\rm old}$
\end{algorithmic}
\end{algorithm}

\subsection{Behavior balancing}
\label{sec:balance}

To reduce the test-weighting imbalance when treating all test as equal in SWE tasks, CodeLift aggregates the tests and patches into groups based on their test result behavior.
For each issue $x$, we evaluate $n\ge2$ final patches on its $m_x$ golden tests to form $M\in\{0,1\}^{{n\times m_x}}$. Each row records one patch's test outcomes, and each column records one test's outcomes across patches. We group identical rows into patch groups $\mathcal G_{\mathrm{patch}}$ and identical columns into test groups $\mathcal G_{\mathrm{test}}$ (Algorithm~\ref{alg:sbr-training}, lines 4--6).
With $Y_{gp}$ denoting test group $g$'s binary outcome on patch group $p$, we compute its failure frequency $d_g$ and normalized weight $w_g$:
\begin{equation}
  d_g=\frac{1}{|\mathcal G_{\mathrm{patch}}|}\sum_{p\in\mathcal G_{\mathrm{patch}}}(1-Y_{gp}),
  \qquad w_g=\frac{d_g}{\sum_{h\in\mathcal G_{\mathrm{test}}}d_h}.
  \label{eq:weights}
\end{equation}
where $\sum_g w_g=1$, or $w_g=0$ if all $d_g=0$,
and CodeLift's final-patch rewards always lie in $[0,1]$.
By assigning one weight per test group, we reduce the influence of repeated tests that might cover the same functionality across P2P tests and F2P tests.
This also automatically remove the influence of those trivial all-passing tests, and gives more weight to groups passed by fewer patch groups, focusing on what is still challenging currently.

Under such behavior balancing mechanism, the test results matrix and corresponding groups and weights updates as the policy evolves, which naturally introduces an adaptive curriculum.
For example, as new patches pass some previously failed tests, such test might be merged to an all-pass groups with zero-weight since the cannot distinguish patch quality, where others might still has weights.
Instead, if one patch failed on a test which is from a previous all-pass test group (i.e., regression), this test group will again obtain a positive weight when other patches still pass it.

\subsection{Snapshot slicing}
\label{sec:slicing}

Before assign credits for intermediate steps, we need to first scoring the intermediate steps.
CodeLift uses self-tests as checkpoints to cut traces into token intervals.
The intuition is that when self-testing, the agent may be checking whether its current edits work and can be used as a candidate submission.
In practice, if the rollout ended there, its cumulative edits can produce a code diff snapshot just like a candidate final patch.
Thus, we can evaluate those snapshots with the same golden test suites and scoring rule (use same $\mathcal G_{\mathrm{test}}$, $\mathcal G_{\mathrm{patch}}$, $w_g$ as final score).

For rollout $j$, we cut $K_j$ slices with token intervals $I_i^j$ and cumulative endpoint states $s_i^j$. Starting from the original state $s_0^j$, we keep a boundary only when the complete golden-test vector changes, so changes with equal scalar scores remain distinct. We assign actions to the slice ending at the next change, with trailing actions in the final slice. If no vector changes, the whole rollout forms one slice. Each token $k\in I_i^j$ receives $A_i^j$, with earlier history as context. Without self-tests, we skip slicing and use the final advantage throughout.

We reuse behavior balancing's groups and weights, keeping every slice reward on the same $[0,1]$ scale. With these fixed for the issue, we compute the state score $S(s)$ as
\begin{equation}
  {S(s)=\sum_{g\in\mathcal G_{\mathrm{test}}}\frac{w_g}{|g|}\sum_{t\in g}o_t(s).}
  \label{eq:score}
\end{equation}
We average the $|g|$ test outcomes within each group because tests that agree on final patches can disagree at intermediate states.

\subsection{Credit assignment}
\label{sec:advantage}

We use these intermediate outcomes to refine credit within each rollout
while preserving the comparison between final patches. Since snapshots
and final patches share a scoring rule, we add their score difference
to the final advantage $A_{\mathrm{final}}^j$; $\Delta_i^j$ denotes this snapshot correction, with $r_{\mathrm{final}}^j=S(s_{\mathrm{final}}(\tau^j))$:
\begin{equation}
  \Delta_i^j=S(s_i^j)-{r_{\mathrm{final}}^j},
  \qquad A_i^j=\Delta_i^j+A_{\mathrm{final}}^j.
  \label{eq:general-advantage}
\end{equation}
CodeLift thus gives an earlier patch with a higher score more credit even if
later edits lower the score (Figure~\ref{fig:overview}).
For the final slice, it keeps $A_{\mathrm{final}}^j$ because the correction is
zero; a single slice recovers final-score training. A positive correction
increases credit, although the resulting advantage can still be negative.

Under our default leave-one-out (LOO) estimator~\citep{ahmadian2024rloo}, this simplifies to
\begin{equation}
  b_j=\frac{1}{{n}-1}\sum_{j'\ne j}{r_{\mathrm{final}}^{j'}},
  \qquad A_{\mathrm{final}}^j={r_{\mathrm{final}}^j}-b_j,
  \qquad A_i^j=S(s_i^j)-b_j.
  \label{eq:slice-rloo}
\end{equation}
We therefore compare each endpoint with the other $n-1$ rollouts' mean
final score, giving positive credit when it exceeds that reference. Here, we
include repeated patch behaviors in the mean because each rollout is a
separate sample.
The fitted scoring rule still depends on rollout $j$'s
outcomes, so we optimize a state-score surrogate for final success, holding
scores and credits fixed.
We provide the policy loss and alternative estimator in
Appendix~\ref{app:implementation}, and work through Figure~\ref{fig:overview}
in Appendix~\ref{app:worked-example}.

\section{Experiments}
\label{sec:experiments}

\subsection{Setup}

\paragraph{Settings}
Our main comparison and ablation settings use Qwen3-30B-A3B-Instruct-2507~\citep{qwen3technicalreport} with R2E-Gym \citep{jain2025r2egym}.
For robustness analysis, we further add Qwen3-Coder-30B-A3B-Instruct (Qwen3-Coder) with SWE-rebench v1 \citep{badertdinov2025rebench}.
We use 500 SWE-bench Verified and 266 Python SWE-bench Pro issues
\citep{deng2025swebenchpro} for evaluation, where Verified closer to \modify{mostly} single-file
training tasks \modify{(R2E-Gym)} and Pro focusing more complex multi-file changes.
We report unbiased pass@$k$ uses eight attempts per issue
\citep{chen2021code,li2026exploring}.
All bases on SWE-agent~\citep{yang2024swe}.
Appendices~\ref{app:implementation}--\ref{app:evaluation} provides more details.

\paragraph{Baselines}
We mainly compare CodeLift with Binary and PassRate (Equations~\ref{eq:success}--\ref{eq:pass-rate-equation}).
We also compare \modify{four} enhanced pass-rate (PR) variants:
(1) \textit{PR-Harmonic} takes the harmonic mean of F2P and P2P pass rates to mitigate progress-regression imbalance.
(2) \textit{PR-Rewighted} uses the static test weights estimated via pass-rate before training~\citep{xiaomi2025mimo,li2026exploring} to mitigate difficulty imbalance.
\modify{
(3) \textit{PR-VeRPO} uses the dynamic test weights estimated via each rollout group's test pass rates and kernel density for training dynamics and process credit assignment~\citep{wang2026beyond}.
(4) \textit{PR-Process} is a variant of PR-Harmonic but uses the same snapshot slicing mechanism as CodeLift.
For ablation, we use a sequential removal for CodeLift's major components: \textit{behavior balancing} and \textit{snapshot slicing}.
We provide \textit{CL-Outcome}, which removes snapshot slicing from CodeLift and assigns final credit to every token.
We then replace behavior-balanced scoring with a binary reward to obtain \textit{Binary}.
Note that here we use the sequential ablation to evaluate snapshot slicing under a fixed dense scoring rule, since binary scores cannot distinguish partial progress among snapshots that do not pass all tests.
}
We aslo compare the default LOO advantage estimator (RLOO~\citep{ahmadian2024rloo}) with group-relative Z-score estimator (GRPO~\citep{shao2024deepseekmath}) in Qwen3-Coder.
More details in Appendix~\ref{app:ablations}.
All share the same setting unless specified.

\subsection{Main comparison}

\begin{table}[!htb]
\centering
\setlength{\tabcolsep}{3pt}
\caption{\textbf{Main results on SWE benchmarks.} Entries are pass@$k$ (\%); Base is the initial model. Deltas show CodeLift gains in percentage points, and bold marks the highest score.
We report raw \textit{PassRate} and \textit{PR-Reweighted} through their final evaluated checkpoint (step 50) in Figure~\ref{fig:validation}b.}
\label{tab:main}
\begin{tabular}{lrrrrrrrr}
\toprule
\textbf{Method} & \multicolumn{4}{c}{\textbf{SWE-bench Verified}} & \multicolumn{4}{c}{\textbf{SWE-bench Pro (Py)}} \\
\cmidrule(lr){2-5}\cmidrule(lr){6-9}
 & pass@1 & pass@2 & pass@4 & pass@8 & pass@1 & pass@2 & pass@4 & pass@8 \\
\midrule
Base & 26.00 & 35.14 & 43.52 & 50.60 & 12.17 & 18.96 & 27.00 & 35.71 \\
Binary & 34.63 & 43.04 & 50.36 & 56.80 & 16.54 & 25.00 & 34.16 & 41.73 \\
PR-Harmonic & 34.45 & 43.76 & 51.62 & 58.00 & 16.92 & 24.70 & 32.56 & 38.72 \\
\modify{PR-VeRPO} & \modify{34.45} & \modify{42.41} & \modify{49.43} & \modify{55.40} & \modify{19.50} & \modify{28.24} & \modify{36.62} & \modify{42.86} \\
\modify{PR-Process} & \modify{35.55} & \modify{44.11} & \modify{51.60} & \modify{58.40} & \modify{19.36} & \modify{27.48} & \modify{35.79} & \modify{42.48} \\
\textbf{CodeLift} & \textbf{38.30} & \textbf{47.53} & \textbf{55.84} & \textbf{62.80} & \textbf{20.16} & \textbf{28.71} & \textbf{36.84} & \textbf{43.23} \\
\midrule
$\Delta$ vs. Binary & \gain{+3.67} & \gain{+4.49} & \gain{+5.48} & \gain{+6.00} & \gain{+3.62} & \gain{+3.71} & \gain{+2.68} & \gain{+1.50} \\
$\Delta$ vs. PR-Harmonic & \gain{+3.85} & \gain{+3.77} & \gain{+4.22} & \gain{+4.80} & \gain{+3.24} & \gain{+4.01} & \gain{+4.28} & \gain{+4.51} \\
\modify{$\Delta$ vs. PR-VeRPO} & \gain{+3.85} & \gain{+5.12} & \gain{+6.41} & \gain{+7.40} & \gain{+0.66} & \gain{+0.47} & \gain{+0.22} & \gain{+0.37} \\
\modify{$\Delta$ vs. PR-Process} & \gain{+2.75} & \gain{+3.42} & \gain{+4.24} & \gain{+4.40} & \gain{+0.80} & \gain{+1.23} & \gain{+1.05} & \gain{+0.75} \\
\bottomrule
\end{tabular}
\end{table}

\paragraph{CodeLift consistently improves performance and broadens issue coverage.}
Table~\ref{tab:main} shows that CodeLift outperforms others on all benchs at every shared reported sampling budget.
On Verified, it outperforms Binary from 3.67 points at pass@1 to 6.00 at pass@8.
We also conducted 95\% paired issue-bootstrap intervals to support CodeLift's advantage (Appendix~\ref{app:uncertainty}).
Results on Pro further suggest that CodeLift \textit{generalizes} to more complex multi-file tasks though trained on single-file tasks.
\modify{When comparing with our enhanced pass-rate baselines (PR-Harmonic \& PR-Process) and adapted reward baselines from existing competitive programming methods (PR-Reweighted \& PR-VeRPO)~\citep{xiaomi2025mimo,li2026exploring,wang2026beyond}, CodeLift still shows an advantage across benchmarks and settings.}

\begin{figure}[!htb]
\centering
\includegraphics[width=\textwidth]{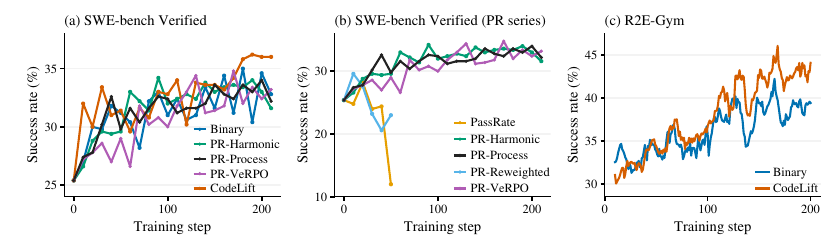}
\caption[Validation and training success rates]{\textbf{Training dynamics.}
Scores monitored during training: (a,b) SWE-bench Verified scores (every 10 steps) and (c) R2E-Gym training rollouts (10-step moving average).}
\label{fig:validation}
\end{figure}

\paragraph{CodeLift sustains learning where pass-rate rewards degrade.}
Figure~\ref{fig:validation} shows that the dense signals of pass-rate rewards do not raise the performance celling of binary rewards in SWE:
The raw PassRate collapses very fast, and PR-Reweighted also shows a clear drop.
While \modify{PR-Harmonic and PR-VeRPO avoid} collapse, their final performance remains comparable to Binary, which aligns with prior findings in competitive programming~\citep{li2026exploring}\modify{, no matter it is static weighting~\citep{xiaomi2025mimo}, dynamic weighting~\citep{wang2026beyond}, or solely mitigating the P2P/F2P imbalance.
Though snapshot slicing improves PR-Process's performance over PR-Harmonic in Table~\ref{tab:main}, it is still sub-optimal to CodeLift.
}
CodeLift also show a stable performance at the ending of training process, staying at 35.8--36.2\% after steps 180, compared with Binary's 30.4--35.0\%.
We further analyze such collapses of pass-rate baselines and stability of CodeLift in Section~\ref{sec:behavior-analysis}.

\begin{table}[!htb]
\begin{minipage}[t]{0.48\textwidth}

\centering
\small
\setlength{\tabcolsep}{1.2pt}
\captionof{table}{\textbf{Ablation for CodeLift's components.}
CL-Outcome removes snapshot slicing; Binary further removes behavior-balanced scoring.
}
\label{tab:snapshot}
\begin{tabular}{lrrrr}
\toprule
\textbf{Method} & \multicolumn{2}{c}{\textbf{Verified}} & \multicolumn{2}{c}{\textbf{Pro (Py)}} \\
\cmidrule(lr){2-3}\cmidrule(lr){4-5}
 & pass@1 & pass@4 & pass@1 & pass@4 \\
\midrule
CodeLift & 38.30 & 55.84 & 20.16 & 36.84 \\
\midrule
{$-\Delta$ snapshot} & \loss{-2.47} & \loss{-4.47} & \loss{-0.28} & \loss{-0.15} \\
{CL-Outcome} & 35.83 & 51.37 & 19.88 & 36.69 \\
$-\Delta$ behavior & \loss{-1.20} & \loss{-1.01} & \loss{-3.34} & \loss{-2.53} \\
Binary & 34.63 & 50.36 & 16.54 & 34.16 \\
\bottomrule
\end{tabular}

\end{minipage}\hfill
\begin{minipage}[t]{0.48\textwidth}

\centering
\small
\setlength{\tabcolsep}{1.2pt}
\captionof{table}{\textbf{Robustness across training settings.} Qwen3-Coder trained on SWE-rebench with RLOO or GRPO.
Base is the initial model.}
\label{tab:sensitivity}
\begin{tabular}{llrrrr}
\toprule
\textbf{Method} & \textbf{Adv.} & \multicolumn{2}{c}{\textbf{Verified}} & \multicolumn{2}{c}{\textbf{Pro (Py)}} \\
\cmidrule(lr){3-4}\cmidrule(lr){5-6}
 & & pass@1 & pass@4 & pass@1 & pass@4 \\
\midrule
Base & $-$ & 44.13 & 57.80 & 34.59 & 49.94 \\
\midrule
Binary & RLOO & 47.70 & 63.20 & 34.30 & 50.88 \\
Binary & GRPO & 47.05 & 61.36 & 34.77 & 50.38 \\
CodeLift & RLOO & 48.30 & 65.40& 35.95 & 51.76 \\
CodeLift & GRPO & 48.80 & 65.80 & 35.76 & 51.40 \\
\bottomrule
\end{tabular}

\end{minipage}
\end{table}%

\subsection{Ablation Study and Robustness Analysis}

\paragraph{CodeLift benefits from both behavior balancing and snapshot slicing.}

Table~\ref{tab:snapshot} shows lower scores when each component is removed step-by-step.
We find that behavior-balanced scoring impacts more on Pro, while snapshot slicing impacts more on Verified. %
Sepcifically, behavior-balanced scoring improves Pro pass@1 over Binary by 3.34pp %
and snapshot slicing improves Verified pass@1 over CL-Outcome by 2.47pp.
\modify{
We additionally compare PR-Harmonic and PR-Process in Table~\ref{tab:main} to assess snapshot slicing under an alternative dense scoring rule as an extra proxy reference.
}

\paragraph{CodeLift improves performance across training settings.}
Table~\ref{tab:sensitivity} shows that CodeLift outperforms Binary on both benchmarks with either RLOO's or GRPO's advantage estimator in the Qwen3-Coder/SWE-rebench setting.
Within each methods, the advantage estimators make little difference on their performance.
On Verified, CodeLift improves pass@4 by 2.20 points with RLOO and 4.44 points with GRPO, supporting broader issue coverage under both estimators.
\section{Analysis}
\label{sec:analysis}
\begingroup
\setlength{\abovecaptionskip}{3pt}
\setlength{\intextsep}{3pt}
\setlength{\columnsep}{6pt}
\makeatletter
\newcommand{\analysisfinishwrap}{\par
  \ifnum\c@WF@wrappedlines>1
    \dimen0=\c@WF@wrappedlines\baselineskip
    \advance\dimen0 by -\baselineskip\vskip\dimen0
  \fi\WFclear}
\newcommand{\analysisneedspace}[1]{\analysisfinishwrap
  \penalty0\relax 
  \dimen0=\pagegoal\advance\dimen0 by -\pagetotal
  \ifdim\dimen0<#1\relax\newpage\fi}
\makeatother

\analysisneedspace{124bp}
\begin{wrapfigure}[11]{r}{166.32bp}
  \centering
  \includegraphics[width=\linewidth]{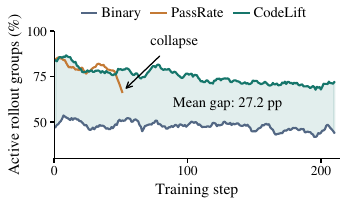}
  \caption{
  \raggedright \textbf{Training signal density}: rollout groups with nonzero advantages (\%).}
  \label{fig:signal-density}
\end{wrapfigure}%

\paragraph{CodeLift provides dense and stable signals.}
\label{sec:signal-density}
Figure~\ref{fig:signal-density} shows that CodeLift consistently provides more rollout groups with non-zero advantages than binary reward throughout the training, as the binary reward gives no relative signal when all sampled patches fail, even if some pass more tests than others.
While pass-rate rewards can distinguish these partially correct patches, their dense feedback does not ensure stable learning and collapses in the early training stage.
CodeLift maintains dense training signals throughout training process, %
with a mean of 75.1\% active rollout groups, outperforming Binary's 47.9\%.
More details are in Appendix~\ref{app:signal-containment}.

\analysisneedspace{195bp}
\begin{wrapfigure}{r}{174.24bp}
  \centering
  \includegraphics[width=\linewidth]{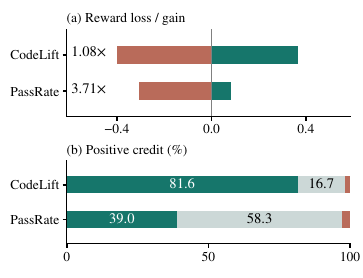}
  \caption{
  \raggedright \textbf{\modify{Progress-Regression preferences}}: (a) Ratio of regression penalty to progress reward.
(b) Positive credit by final-patch test pass rate: improved (green), unchanged (gray), worsened (red).}
  \label{fig:fix-regression}
\end{wrapfigure}%

\paragraph{Behavior balancing reduces pass-rate bias toward overly conservative.}
\label{sec:behavior-analysis}
To understand why PassRate collapses while CodeLift remains stable, we (1) compare rewards for \textit{progress} and \textit{regression} and (2)
which types of patches receive positive training signals.
As shown in Figure~\ref{fig:fix-regression}, we find that with raw PassRate, \textit{when final patches make any changes, the mean final penalty for introducing regressions is \textbf{3.71} times than the mean final reward for making a real progress.}
Thus, the model is encouraged to be \textit{conservative}, i.e., the potential reward for making progress may not be worth the risk of a large penalty, with \textbf{58.3\%} positive advantages come from the patches without pass rate changes.
CodeLift mitigates such bias by balancing the reward contribution, avoiding the tests with similar behavior dominates the reward.
As a result, it achieves a 1.08 ratio, close to parity.
This is also reflected in its allocation of positive advantages, as 81.6\% of them come from patches that make real progress.

\analysisneedspace{120bp}
\begin{wrapfigure}[11]{r}{166.32bp}
  \centering
  \includegraphics[width=\linewidth]{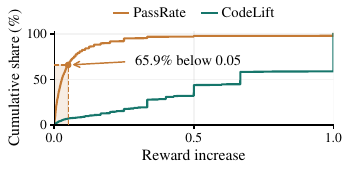}
  \caption{
  \raggedright Positive reward changes for fully passing a test family.}
  \label{fig:property-contribution}
\end{wrapfigure}%

We further examine how test counts per parameterized test familes contribute to this reward imbalance.
Intuitvely, a
functionality
covered by few tests have little impacts on pass-rate reward even when all its tests pass.
Figure~\ref{fig:property-contribution} compares positive reward changes for patches gaining exactly one fully passing test family. Under PassRate, 65.9\% of these changes are below 0.05.
Such small reward increases are less common under CodeLift.
More details are in Appendix~\ref{app:positive-credit}.

\analysisneedspace{190bp}
\begin{figure}[H]
  \centering
  \includegraphics[width=\linewidth]{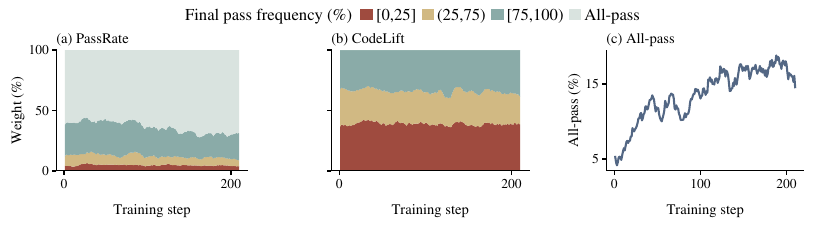}
  \caption{\raggedright \textbf{Test weights \modify{training dynamics}}.
  \modify{(a,b) Test-weight shares within rollout groups under PassRate and CodeLift.
  (c) Share of rollout groups where every final patch passes all tests.}}
  \label{fig:difficulty}
\end{figure}%

\paragraph{CodeLift naturally introduces an adaptive curriculum.}
\label{sec:curriculum}
To see how the test weights change as the policy improves, we compare PassRate and CodeLift on the same training batches.
As shown in Figure~\ref{fig:difficulty}, we sort tests by how often sampled patches pass them.
\textit{\modify{Under PassRate, most of the total weight is assigned to tests passed by every patch.}}
Thus, most of its reward scale does not distinguish the sampled solutions, and the space for challenging test's weights gradually decreases.
CodeLift combines these trivial tests into one group with zero weight, and then assigns more weight to groups that fewer distinct patch groups pass.
As more tests pass on every sampled patch, they join the all-pass group
without occupying weight spaces.
\textit{\modify{Thus, the weight distribution dynamically keeps \textit{stable} throughout the training.}}

We further track test-group counts and weights to examine how the curriculum changes as the policy improves.
As shown in Figure~\ref{fig:class-dynamics}, easy test groups roughly remain stable in both high counts and carry low average weights.
\analysisneedspace{170bp}
\begin{wrapfigure}[16]{r}{217.80bp}
  \centering
  \includegraphics[width=\linewidth]{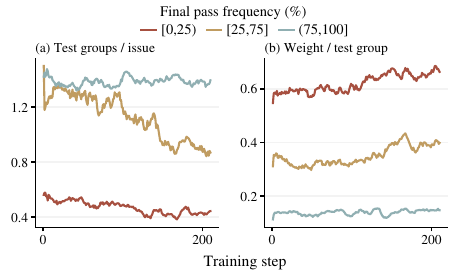}
  \caption{
  \raggedright \textbf{Test groups \modify{training dynamics}}. (a) Groups per issue. (b) Average weight per group.}
  \label{fig:class-dynamics}
\end{wrapfigure}%

\noindent This follows the grouping mechanism: simple tests universally passed by patches will gradually join the same all-pass groups without adding weight.
Medium-difficulty groups show the largest decline in number, consistent with partially learned tests becoming easier to pass.
Hard groups decrease much more slowly, but their average weight keeps rising, so passing one hard group contributes more to the reward on average.
Our analysis of the same issue's test groups across training windows further shows that their weights change with the policy's performance:
Comparing with previous windows, CodeLift gives less weight when more patches pass its tests, and more weight when fewer pass on average (Appendix~\ref{app:matched-curriculum}).

\analysisneedspace{182bp}
\begin{wrapfigure}[16]{r}{217.80bp}
  \centering
  \includegraphics[width=\linewidth]{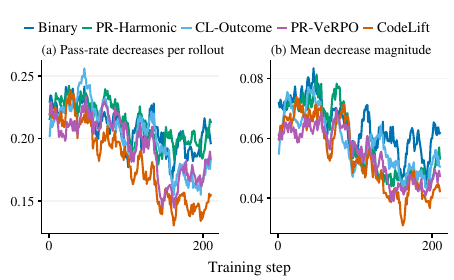}
  \caption{
  \raggedright \textbf{Regressions \modify{training dynamics}}. (a) Pass-rate decreases \modify{\textit{count}}. (b) decrease \textit{magnitude}.}
  \label{fig:process}
\end{wrapfigure}%

\paragraph{Snapshot slicing improves credit assignment and reduces regressions.}
\label{sec:process}
We track how often %
test pass rates fall between consecutive slice endpoints and the size of these decreases.
As shown in Figure~\ref{fig:process}, CodeLift has fewer decreases per rollout
(a) and smaller mean decreases (b) than Binary or CL-Outcome late in training.
To check whether final-only credit misses useful intermediate work, we compare slice advantages with and without slicing on the same CodeLift rollouts, keeping state scores and the peer baseline fixed.
Final-only credit gives every slice the final patch's advantage, even when an intermediate patch scores higher.
With slicing, CodeLift changes zero or negative advantages into positive ones in 23.4\% of intermediate slices, providing credit to useful actions within these rollouts (Appendix~\ref{app:snapshot-credit}).
This gives credit to intermediate work even when later edits lower the final score.

\paragraph{CodeLift's Curriculum does not need extra rollout effort.}
By design, CodeLift's behavior balancing estimates test difficulty from the final-patch test results already collected during normal test-based RL training.
As the policy improves, these results change and the groups and weights adapt, naturally introducing a curriculum without extra rollouts.
In contrast, similar pass-rate rewards for competitive programming~\citep{xiaomi2025mimo,li2026exploring} use separately sampled offline test pass frequencies for offline test difficulty estimation.
However, CodeLift does need extra testing effort for snapshot slicing, which evaluates the intermediate snapshot's patches and provides extra credit assignments.
In our main CodeLift experiment, snapshot slicing requires evaluating an average of 2.35 additional non-final patches per rollout beyond the final patch.
These additional test executions run on CPUs rather than GPUs and require no additional rollout generation.

\analysisfinishwrap
\endgroup

\section{Related Work}

\paragraph{Repository-level SWE tasks.}

SWE-bench evaluates repository issue resolution~\citep{jimenez2024swebench},
while SWE-agent provides navigation, editing, and execution
interfaces~\citep{yang2024swe}.
R2E-Gym and SWE-rebench supply executable training
tasks~\citep{jain2025r2egym,badertdinov2025rebench}.
R2E-Gym fine-tunes agents on successful trajectories.
CodeLift builds on these datasets and environment, extending final-outcome
RL with behavior-balanced rewards and intermediate credit.

\paragraph{\modify{Coding rewards.}}

MiMo~\citep{xiaomi2025mimo} calibrates test difficulty using multiple models' solutions and assigns partial rewards by difficulty level.
\citet{li2026exploring} find no consistent final-performance gains from raw or statically reweighted pass-rate rewards over binary rewards.
VeRPO~\citep{wang2026beyond} weights tests by current pass frequencies and their density, combining per-turn and final rewards.
These methods use \textit{test frequency statistics} for \textit{competitive programming}.
In contrast, \textit{repository-level SWE} additionally requires preserving existing functionality (P2P) while implementing requested changes (F2P).
\modify{CodeLift exploits the more fundamental patch \textit{testing behavior mechanism} to provide a balanced, dense supervision, with grouped tests as dynamic curriculum.}
\modify{To fairly compare with related works in SWE scenarios,} our PR-Reweighted baseline further adapts the difficulty-driven static calibration philosophy (i.e., \textit{IOI partial scoring}) from MiMo and
\citet{li2026exploring}, \modify{and dynamic pass-rate frequency density weighting from VeRPO~\citep{wang2026beyond}} to SWE.
We also introduce PR-Harmonic \modify{and PR-Process} to balance F2P/P2P pass rates \modify{with or without our proposed snapshot slicing credit assignment}, which \modify{are} enhanced pass-rate \modify{baselines} specific for SWE tasks (Section~\ref{sec:experiments}).

\paragraph{Fine-grained credit assignment.}

GiGPO~\citep{feng2025gigpo} compares subsequent returns from actions at repeated states.
TreeRL~\citep{hou2025treerl} and Branching Policy Optimization (BPO)~\citep{he2026bpo} estimate intermediate values from
sampled continuations.
CodeLift instead evaluates the patch already produced at each checkpoint and
uses its test score to credit the preceding interaction slice.

\section{Limitations}
\label{sec:limitations}

We train and evaluate CodeLift on Python repositories, using established
public datasets with executable tests. These resources let us focus on
the RL method; building comparable datasets for other languages is
outside the scope of this work. Accordingly, our self-test detection,
and data checks have been validated on Python projects
with test runners such as \texttt{pytest}. CodeLift's scoring and credit
rules use test outcomes, but its performance in other languages remains
to be evaluated.

\section{Conclusion}

We introduce CodeLift, which combines behavior balancing and snapshot slicing to provide dense rewards for SWE agents.
Its behavior balancing groups tests and patches to reduce pass-rate imbalances, which naturally introduces an adaptive curriculum from existing final-patch outcomes.
Its snapshot slicing applies the same scoring rule to intermediate patches, assigning credit for steps with progress even when later edits introduce regressions.
Experiments on SWE-bench Verified and SWE-bench Pro (Py) show consistent advantage over binary and enhanced pass-rate baselines.
Our analyses show denser learning signals, more balanced rewards, and test weights that adapt to the changing policy.

\clearpage






\paragraph{Reproducibility statement.}
\modify{Code repository is in \url{https://github.com/Siyuexi/CodeLift}}

\bibliography{sbr_references}
\bibliographystyle{iclr2027_conference}

\clearpage
\appendix
\section{Experimental Details}
\label{app:implementation}

\subsection{Setting details}
\begin{wraptable}{r}{0.50\textwidth}
  \centering
\caption{Shared training parameters.}
\label{tab:training-settings}
\begin{tabular}{lr}
\toprule
Parameter & Value \\
\midrule
Issues per batch & 64 \\
Rollouts per issue & 8 \\
Minibatch (issues) & 16 \\
Update epochs & 1 \\
Learning rate & $10^{-6}$ \\
Clip width & $4\times10^{-4}$ \\
Prompt tokens & 8,192 \\
Response tokens & 122,880 \\
Training temperature & 1.0 \\
Policy loss & GSPO \\
Advantage estimator & RLOO \\
Entropy / KL coefficient & 0 / 0 \\
\bottomrule
\end{tabular}\space\space
\end{wraptable}%

We use R2E-Gym-Subset and the filtered SWE-rebench v1 release to train Qwen3 and Qwen3-Coder, respectively.
In our training settings as shown in Table~\ref{tab:training-settings}, we exclude prompts above 8,192 tokens, counting system and issue messages with the model's tokenizer and chat template.
After filtering, all methods share the filtered 4,455 R2E-Gym or 1,100 SWE-rebench issues.

We use the SWE-agent harness for training and evaluation, which provides a persistent shell, a string-replacement editor, and a submit action.
The agent can inspect and run existing repository tests, while
additional benchmark tests and grading scripts remain in the evaluator.
For both training and evaluation, we set a 120-second shell-command timeout.
During training, we also shuffle issues and set the dataset and rollout seeds to 42.

For fair comparison, we fix all training seeds to 42, and 8-repeat evaluation with different seeds: 42, 4, 424, 2424, 44, 45, 46, 47.
All the settings for training and evaluation, including the infrastructure, deployment, framework, datasets, seeds, algorithm, keep consistent across all methods to ensure fairness.
The only difference among the methods is their own reward calculation and credit assignment.

\subsection{Reward hacking prevention}
To prevent model reward hacking during all method's training and evaluation settings, we limit model's access to reference solutions by we disabling network access in the sandbox and replace the repository's Git history with a single initial commit and no remotes. These safeguards apply to all compared methods.

\subsection{Engineering Details}
CodeLift captures cumulative patches after recognized self-tests
and at the end of each rollout. For self-tests, we recognize Python test
runners (\texttt{pytest}, \texttt{unittest}, \texttt{tox}, and \texttt{nox}),
\texttt{make test}/\texttt{make check}, and test shell scripts, capturing
the patch once the command finishes regardless of its exit status.

It is important to note that self-test outputs do not determine reward or enter the weight calculation in any situation.
Self-test commands select candidate checkpoints, which we evaluate on the same golden tests as the original and final states. These test vectors determine slice boundaries and scores (Section~\ref{sec:slicing}).

\subsection{Optimization and advantage estimators}
We use GSPO to compute the likelihood ratio and clipped policy loss~\citep{zheng2025gspo}, with RLOO~\citep{ahmadian2024rloo} or GRPO~\citep{shao2024deepseekmath} for advantage estimation.
For token $k$ in rollout $j$, let $p_{jk}(\theta)$ be its probability given the
history and $L_j$ the number of generated tokens. GSPO uses a sequence ratio $\rho_j$ and its token form $\rho_{jk}$ for our slice advantages:
\begin{align}
 {\rho_j(\theta)}
 &{=\exp\!\left(\frac{1}{L_j}\sum_{k=1}^{L_j}\log\frac{p_{jk}(\theta)}{p_{jk}(\theta_{\mathrm{old}})}\right),}\\
 {\rho_{jk}(\theta)}
 &{=\sg[\rho_j(\theta)]\,\frac{p_{jk}(\theta)}{\sg[p_{jk}(\theta)]}.}
\end{align}
Here $\sg$ preserves the forward value and stops its gradient.
Let $\widehat A_k^j=A_i^j$ for $k\in I_i^j$, or $\widehat A_k^j=A_{\mathrm{final}}^j$ when slicing is skipped. We use
\begin{equation}
 \mathcal L_j(\theta)=-\frac{1}{L_j}\sum_{k=1}^{L_j}
 \min\!\left\{\rho_{jk}(\theta){\widehat A_k^j},
 \operatorname{clip}(\rho_{jk}(\theta),1-\epsilon,1+\epsilon){\widehat A_k^j}\right\}.
 \label{eq:policy-loss}
\end{equation}
We then average the loss over tokens within each rollout and over
rollouts in the minibatch, with $\epsilon=4\times10^{-4}$. Each update uses
one epoch and no entropy or KL regularization.

By default, we use REINFORCE Leave-One-Out (RLOO. Equation~\ref{eq:slice-rloo}).
For the GRPO comparison, we change only the final-advantage estimator:
let $\mu$ and $\sigma$ be the mean and sample standard deviation of the
issue group's final rewards. We compute the final advantage as
\begin{equation}
 A_{\mathrm{final}}^j=\frac{r_{\mathrm{final}}^j-\mu}{\sigma+10^{-6}},
\end{equation}
where $r_{\mathrm{final}}^j$ is the Binary reward or CodeLift score.
CodeLift then adds the snapshot-to-final score difference at its
original scale (Equation~\ref{eq:general-advantage}), so only the final
comparison is standardized.

\subsection{Reward baselines and ablation}
\label{app:ablations}
\paragraph{PR-Harmonic.}
We use the harmonic mean value of P2P pass rate and F2P pass rate to calibrate their imbalance.
Let $f(s)$ and $p(s)$ be the pass rates on the original F2P and P2P
test sets. When both sets are nonempty, we compute the reward as
\begin{equation}
 R_{\mathrm{PR\text{-}Harmonic}}(s)=\frac{2f(s)p(s)}{f(s)+p(s)}.
\end{equation}
The reward is zero when both rates are zero. If one set is empty,
we use the other set's pass rate. Within each set, tests receive equal weight.

\paragraph{\modify{PR-Process.}}
\modify{We apply PR-Harmonic's score to each repository snapshot and use
CodeLift's snapshot slicing to construct an enhanced pass-rate baseline with credit assignment. This comparison keeps
the harmonic scoring rule and adds credit for intermediate progress.}

\paragraph{PR-Reweighted.}
PR-Reweighted is an SWE-task adaptation for the \modify{test's static weighting} philosophy of existing competitive programming pass-rate baselines~\citep{xiaomi2025mimo} and~\citep{li2026exploring}.
Specifically,
we adapt the test-difficulty labeling procedure of
\citet[Appendix~A, Algorithm~1]{li2026exploring}, inspired by
MiMo's difficulty-based code reward~\citep{xiaomi2025mimo}.
Before training, we estimate test difficulty from final patches generated
by Binary over four successive calibration stages. Let $p_{te}$ be test $t$'s pass frequency in stage $e$,
and $a_e$ that stage's overall success rate. We compute
\begin{equation}
 \widetilde p_t=\frac{\sum_{e\in E_t}a_ep_{te}}{\sum_{e\in E_t}a_e},
 \qquad R_{\mathrm{PR\text{-}Reweighted}}(s)=\frac{\sum_t v_to_t(s)}{\sum_t v_t},
\end{equation}
where $E_t$ contains stages with an observed outcome for test $t$.
Using these estimates, we set the weights $v_t$ before training
and keep them fixed. Specifically, within each issue, we assign weight 1
to tests at or above the 67th percentile of $\widetilde p_t$, weight 3 to
those below the 33rd percentile, and weight 2 to the rest, using linear
interpolation for the percentiles. If all calibrated frequencies are at
least 0.9 or at most 0.1, we assign weight 1 or 3, respectively.
As a fallback, we use weight 2 for tests without positive calibration
weight and for all tests in issues with fewer than eight usable
calibration patches.

\paragraph{\modify{PR-VeRPO.}}
\modify{
PR-VeRPO is an SWE-task adaptation for the test's dynamic weighting via test frequency density method, following the original VeRPO~\citep{wang2026beyond} from its original competitive programming scenario to SWE agentic tasks. For each issue, we pool the candidates captured after
self-tests and at rollout completion, retaining repeated outcomes and excluding
the original repository state. Each test's pass frequency $\rho_t$ across these
candidates determines its weight:
\[
 w_t=\frac{\exp(-\alpha\rho_t)}
 {\sum_u\exp\!\left(-\frac{(\rho_t-\rho_u)^2}{2\sigma^2}\right)+\epsilon},
 \qquad \sigma=\tfrac12\operatorname{std}(\{\rho_t\}).
\]
The population standard deviation is over tests. When all frequencies coincide, we use
the equal-frequency limit, where the kernel sum is the number of tests.
We sum the weights of passed tests without weight-sum normalization to obtain
each candidate's local reward and center these rewards over all candidates
for the issue. We multiply each rollout's final full-test success by
$\gamma^{K_j}$, where $K_j$ is its candidate count, then subtract the mean
discounted success across all rollouts for the issue. Candidate credit is
this centered final reward plus $\beta$ times its centered local reward.
We use $\alpha=2$, $\gamma=0.95$, $\beta=1$, and $\epsilon=10^{-8}$,
and assign the resulting credit to the candidate's generated-token interval.}

\paragraph{CL-Outcome.}
CL-Outcome retains CodeLift's behavior-balanced scoring and snapshot evaluation, but assigns $A_{\mathrm{final}}^j$ to
every generated token. Similarly, PR-Harmonic uses the same snapshot
evaluation, with the rollout advantage computed from its
final reward.

\section{Evaluation and Uncertainty}
\label{app:evaluation}

\paragraph{Benchmarks and checkpoints.}
Before we start any experiment, we set a global target with fixed budgets for evaluation: (1) Qwen3/R2E-Gym: uses step200 (near 3 epochs) for main comparison evaluation. (2) Qwen3-Coder/SWE-rebench: uses step100 (near 6 epochs) for robustness evaluation.
We also set an early-stop protocol before training: if any model's validation score remained below the pre-training validation score for 30 consecutive steps, it was deemed to have collapsed.
For collapsed models, we report the validation curves up to early stopping, including the final evaluated step and its score, without conducting further evaluations across multiple random seeds.
Under these rules, we report all successful method's results in Table~\ref{tab:main} and all collapsed model information in Figure~\ref{fig:validation}.

\paragraph{Repeated evaluation.}
We sample eight attempts per issue at temperature 1.0, allowing
1,200 seconds for each rollout and 600 seconds for evaluation.
For $c_i$ successful attempts on issue $i$, we report
\begin{equation}
 \widehat{\mathrm{pass@}k}=\frac{1}{N}\sum_{i=1}^{N}
 \left[1-\frac{\binom{8-c_i}{k}}{\binom{8}{k}}\right],
 \qquad k\in\{1,2,4,8\},
\end{equation}
with a zero numerator when $8-c_i<k$. %
We count missing attempts as failures,
keeping $N=500$ for Verified and $N=266$ for Pro. To report differences,
we subtract the displayed percentages after rounding to two decimals.
We train each Qwen3 method once, then repeat evaluation attempts
to measure variation in the trained model's sampled outputs.

\paragraph{Paired uncertainty.}
\label{app:uncertainty}

\begin{table}[!htb]
\small
\centering
\caption{CodeLift's gains on Verified with 95\% paired bootstrap intervals (pp).}
\label{tab:paired-uncertainty}
\begin{tabular}{lcccc}
\toprule
\textbf{Method} & \textbf{pass@1} & \textbf{pass@2} & \textbf{pass@4} & \textbf{pass@8} \\
\midrule
\textbf{Binary} & \begin{tabular}[c]{@{}c@{}}+3.67{[+2.10,+5.28]}\end{tabular} & \begin{tabular}[c]{@{}c@{}}+4.49{[+2.61,+6.38]}\end{tabular} & \begin{tabular}[c]{@{}c@{}}+5.48{[+3.12,+7.85]}\end{tabular} & \begin{tabular}[c]{@{}c@{}}+6.00{[+2.80,+9.20]}\end{tabular} \\
\textbf{PR-Harmonic} & \begin{tabular}[c]{@{}c@{}}+3.85{[+2.20,+5.55]}\end{tabular} & \begin{tabular}[c]{@{}c@{}}+3.77{[+1.88,+5.70]}\end{tabular} & \begin{tabular}[c]{@{}c@{}}+4.22{[+1.89,+6.63]}\end{tabular} & \begin{tabular}[c]{@{}c@{}}+4.80{[+1.60,+8.00]}\end{tabular} \\
\modify{\textbf{PR-Process}} & \modify{\begin{tabular}[c]{@{}c@{}}+2.75{[+1.23,+4.30]}\end{tabular}} & \modify{\begin{tabular}[c]{@{}c@{}}+3.42{[+1.69,+5.17]}\end{tabular}} & \modify{\begin{tabular}[c]{@{}c@{}}+4.24{[+1.91,+6.55]}\end{tabular}} & \modify{\begin{tabular}[c]{@{}c@{}}+4.40{[+1.00,+7.80]}\end{tabular}} \\
\modify{\textbf{PR-VeRPO}} & \modify{\begin{tabular}[c]{@{}c@{}}+3.85{[+2.25,+5.48]}\end{tabular}} & \modify{\begin{tabular}[c]{@{}c@{}}+5.12{[+3.22,+7.03]}\end{tabular}} & \modify{\begin{tabular}[c]{@{}c@{}}+6.41{[+3.97,+8.92]}\end{tabular}} & \modify{\begin{tabular}[c]{@{}c@{}}+7.40{[+4.00,+10.80]}\end{tabular}} \\
\modify{\textbf{CL-Outcome}} & \modify{\begin{tabular}[c]{@{}c@{}}+2.47{[+0.95,+4.03]}\end{tabular}} & \modify{\begin{tabular}[c]{@{}c@{}}+3.44{[+1.66,+5.25]}\end{tabular}} & \modify{\begin{tabular}[c]{@{}c@{}}+4.47{[+2.22,+6.75]}\end{tabular}} & \modify{\begin{tabular}[c]{@{}c@{}}+5.00{[+2.00,+8.20]}\end{tabular}} \\
\bottomrule
\end{tabular}
\end{table}

We use paired bootstrap intervals to check CodeLift's gains on
Verified (Table~\ref{tab:paired-uncertainty}), finding gains over Binary,
PR-Harmonic, \modify{PR-Process, PR-VeRPO}, and CL-Outcome across pass@1/2/4/8. For this check, we resample
the 500 issues with replacement 20,000 times, keeping all eight attempts
and the method pairings for each selected issue. The 2.5th and 97.5th
percentiles of the resulting mean differences then give the interval
endpoints. Because only issues are resampled, these intervals measure
variation across issues with the trained models and sampled attempts fixed.

We also check the effect of missing attempts while keeping the
benchmark denominator fixed. Even if all seven missing CL-Outcome attempts
succeed and both missing CodeLift attempts fail, CodeLift retains a
2.30-point Verified pass@1 gain.

\section{Further Analysis Details}
\label{app:diagnostic-definitions}
We use Qwen3 training rollouts for the analyses below unless stated otherwise.
To track progress in Figures~\ref{fig:signal-density},
\ref{fig:difficulty}, and~\ref{fig:class-dynamics}, we reconstruct
successive batches of sampled rollouts (sampling cohorts) and label them \emph{Training step}.

\subsection{Final-patch credit and signal density}
\label{app:property-credit}
Let $y_{gj}$ be test group $g$'s final outcome on rollout $j$, and let
$k_g=\sum_{{j=1}}^{{n}} y_{gj}$ count passing rollouts before patch grouping.
We can write this group's contribution to the RLOO advantage as
\begin{equation}
 a_{gj}=w_g\frac{{n} y_{gj}-k_g}{{n}-1},
 \qquad A_{\mathrm{final}}^j=\sum_g a_{gj},
 \qquad\frac{1}{{n}}\sum_j|a_{gj}|=
 \frac{2w_gk_g({n}-k_g)}{{n}({n}-1)}.
 \label{eq:property-credit}
\end{equation}
Thus, CodeLift assigns credit according to both a test group's weight
and how its outcomes differ across patches. If every final patch fails a
group, CodeLift gives it positive weight but cannot use it to distinguish
those patches. It can still credit an intermediate state that passes the
group. When some final patches pass and others fail, the group contributes
nonzero advantages, though contributions from different groups may cancel.

\label{app:signal-containment}
\label{app:signal-details}
\begin{wraptable}{r}{0.48\textwidth}
  \centering
\caption{
Nonzero-advantage rollout groups on the same CodeLift rollouts.}
\label{tab:signal}
\begin{tabular}{lr}
\toprule
Credit rule & Share (\%) \\
\midrule
Binary & 48.07 \\
CL-Outcome & 74.99 \\
CodeLift & 75.02 \\
\bottomrule
\end{tabular}\space\space
\end{wraptable}%

To separate the credit rule from patch generation, we score the
same patches under each rule (Table~\ref{tab:signal}). CodeLift's behavior
balancing accounts for most of the increase in rollout groups with nonzero
advantages. Beyond this increase, snapshot slicing also changes credit
within groups that already have a signal.

CL-Outcome retains a nonzero signal wherever Binary RLOO has one
on the same rollouts, because such a group includes both a fully passing
patch with score one and a failing patch with score below one. CodeLift
also retains this comparison in the final slice. When a group has no
final-only signal, however, slicing adds one only if an intermediate
score differs from the shared final score.

To track signal density during training, we use each method's own
rollouts (Figure~\ref{fig:signal-density}): cohorts 1--210 for CodeLift and
Binary, and 1--51 for PassRate. Within these ranges, we divide the number
of groups with any nonzero advantage by all scored groups. We then smooth
the curves over ten cohorts and pool group counts for the reported means.

\subsection{Reward changes and positive credit}
\label{app:positive-credit}
We compare rewards for progress and regression by scoring the
same CodeLift patches under both reward rules
(Figures~\ref{fig:fix-regression}a and~\ref{fig:property-contribution}).
For this comparison, each unparameterized test forms a singleton family,
and a family passes when all its tests pass. We then compare each final
patch with its evaluated original state and divide the absolute mean
reward change for a net loss of passing families by the mean change for
a net gain. As a further check, we exclude issues with inconsistent
original outcomes: PassRate's loss/gain ratio changes from 3.71 to 3.68,
while CodeLift's remains at 1.08.

To examine the effect of test counts, we select CodeLift patches
that gain exactly one fully passing family over the original repository
(Figure~\ref{fig:property-contribution}). We then score this set under both
rules, using Equation~\ref{eq:score} for CodeLift, and plot the cumulative
share of positive reward changes for each rule.
\modify{Conditioning separately on positive reward changes yields 36,778 samples for PassRate and 20,609 for CodeLift. The reported percentages below 0.05 are computed within these respective subsets.}

We also examine which patches receive positive credit
(Figure~\ref{fig:fix-regression}b), using each method's own complete
eight-rollout groups from cohorts 49--51. For this analysis, we separate
final patches by whether their raw pass rate increases, stays unchanged,
or decreases from the original state. Within each category, we then sum
$\max(A_{\mathrm{final}}^j,0)$ and divide by the total positive advantage.

\subsection{Adaptive curriculum}
\label{app:curriculum-details}
To track how CodeLift allocates weight as the policy improves,
we sort tests by how often the evaluated final patches pass them in each
issue batch (Figure~\ref{fig:difficulty}). For this comparison, we use
batches with at least one failing patch and report the rate of all-pass
rollout groups separately. We also track test-group counts and weights
within pass-frequency intervals of $[0,25)$, $[25,75]$, and $(75,100]$
percent (Figure~\ref{fig:class-dynamics}). To obtain these curves, we
recompute the bins in each batch and pool eleven neighboring cohorts,
averaging counts per issue occurrence and weights per test group.

\label{app:matched-curriculum}
\begin{table}[!htb]
\centering
\caption{Changes in test groups with fixed membership across training windows.}
\label{tab:matched-curriculum}
\begin{tabular}{lrrrr}
\toprule
 & \multicolumn{2}{c}{Failure frequency $d_g$} & \multicolumn{2}{c}{Weight $w_g$} \\
\cmidrule(lr){2-3}\cmidrule(lr){4-5}
Passing rollouts & Before & After & Before & After \\
\midrule
More & 0.609 & 0.439 & 0.460 & 0.434 \\
Fewer & 0.439 & 0.598 & 0.395 & 0.471 \\
Same & 0.475 & 0.475 & 0.347 & 0.361 \\
\bottomrule
\end{tabular}
\end{table}

We also follow individual test groups over time by matching
groups containing the same tests across training windows
(Table~\ref{tab:matched-curriculum}). For each issue, we first select its
earliest complete eight-rollout group in cohorts 1--70, 71--140, and
141--210. We then retain issues evaluated on the same tests in all three
windows and match groups across consecutive windows. Across the resulting
11,216 pairs, CodeLift gives less weight to groups passed by more rollouts
and more weight to those passed by fewer, on average. Here, pass counts
include all rollouts, whereas $d_g$ uses distinct patch groups. Because
CodeLift also normalizes weights across test groups in the same issue,
a group's weight can change even if the same number of rollouts passes it.

\subsection{Snapshot credit and regression measurements}
\label{app:snapshot-credit}
\begin{table}[!htb]
\centering
\caption{Effect of slicing on recorded rollouts (\%).}
\label{tab:snapshot-credit-accounting}
\begin{tabular}{lrr}
\toprule
Rollout source & CodeLift & CL-Outcome \\
\midrule
Intermediate slices: $A_{\mathrm{final}}\leq0$ to $A_i>0$ & 23.4 & 17.1 \\
Intermediate slices: $A_{\mathrm{final}}>0$ to $A_i<0$ & 23.9 & 24.1 \\
\bottomrule
\end{tabular}
\end{table}

We compare credit with and without slicing on each method's
recorded rollouts, keeping state scores and the peer baseline fixed
(Table~\ref{tab:snapshot-credit-accounting}). This comparison covers cohorts
1--210, with 1,047 intermediate slices for CodeLift and 5,741 for CL-Outcome.
With slicing, CodeLift changes zero or negative advantages under
final-only credit into positive ones for 245 (23.4\%) of its intermediate
slices. It also changes some positive advantages to negative ones,
retaining credit for useful intermediate work and removing it from
intermediate regressions.

\label{app:process-details}
We also measure regressions by counting decreases in raw pass rate
between consecutive credited states, starting from the original
repository (Figure~\ref{fig:process}). To measure their size, we average
decreases within each rollout, assigning zero when none occur, then
average across rollouts. We smooth the curves with an eleven-step
centered window.

\section{Case Study}
\label{app:cases}

\modify{We first work through the full calculation, then use recorded cases
to illustrate behavior balancing's correction of test multiplicity bias
and snapshot slicing's credit for intermediate progress and regressions.}

\subsection{Preliminary: Calculation workflow}
\label{app:worked-example}
\modify{We use the five sampled patches in Figure~\ref{fig:overview} to
illustrate how CodeLift computes rewards and slice advantages from test outcomes.}

\modify{Since P1 and P4 pass the same tests, CodeLift places them in the same
patch group, leaving four distinct patch groups. It also groups T3 and T5
together because both pass on every sampled patch. This gives five test
groups, which reject $(3,3,0,2,2)$ patch groups, respectively. CodeLift
normalizes these counts to weights $(0.3,0.3,0,0.2,0.2)$, then sums the
weights of the test groups each patch passes to obtain final scores
$(0.3,0.2,0.2,0.3,0.7)$.}

\modify{To compute the advantage for rollout 5, CodeLift compares its score
with the mean final score of the other rollouts. P1 and P4 still count as
separate samples here, so the peer baseline is
$b_5=(0.3+0.2+0.2+0.3)/4=0.25$. The final advantage is therefore
$0.7-b_5=0.45$.}

\begin{table}[!htb]
\centering
\caption{\modify{\textbf{Slice credit for rollout 5 in Figure~\ref{fig:overview}.}}}
\label{tab:example-credit}
\begin{tabular}{lccc}
\toprule
\modify{\textbf{Endpoint}} & \modify{\shortstack{\textbf{Score} $S(s_i)$}} & \modify{\shortstack{\textbf{Correction} $\Delta_i$}} & \modify{\shortstack{\textbf{Advantage} $A_i$}} \\
\midrule
\modify{\textbf{First snapshot}} & 0.5 & $-0.2$ & $+0.25$ \\
\modify{\textbf{Second snapshot}} & 0.8 & $+0.1$ & $+0.55$ \\
\modify{\textbf{Final patch}} & 0.7 & $0$ & $+0.45$ \\
\bottomrule
\end{tabular}
\end{table}

\modify{CodeLift uses this same baseline to assign credit to each slice
(Table~\ref{tab:example-credit}). For example, the first snapshot passes T2
and T4 and receives a score of 0.5. This is lower than the rollout's final
score but higher than the peer baseline. Its snapshot correction is
therefore negative, while its slice advantage is positive.}

\subsection{\modify{Behavior balancing for test multiplicity}}
\label{app:case-balancing}
\modify{In this subsection, we discuss a pandas issue involving three
internal classes: \texttt{Block}, \texttt{DatetimeTZBlock}, and
\texttt{ExtensionBlock}. Accessing these classes should emit a deprecation
warning, and they should no longer appear in the exposed module namespace.
The existing \texttt{make\_block} function must also remain usable for
constructing blocks from timezone-aware datetime data. The tests check
these three behaviors, but the warning check is repeated for each class,
while the namespace and block construction checks each have one test.
We compare two final patches fron an 8-rollout groups in Table~\ref{tab:case-balancing}. Final patch A adds the required
warnings but leaves the namespace incorrect and breaks block construction.
Final patch B corrects the namespace and preserves block construction,
but still fails the warning checks.}

\begin{table}[!htb]
\centering
\caption{\modify{\textbf{Test outcomes and final rewards for two pandas patches.}
Counts report passed/total tests; $\checkmark$/$\times$ mark passing/failing
test groups. Weights are computed from all eight final patches in the
rollout group. The higher reward under each rule is bold.}}
\label{tab:case-balancing}
\begin{tabular}{lccc}
\toprule
\modify{\textbf{Tested behavior}} & \modify{\textbf{Group weight}} & \modify{\textbf{Final patch A}} & \modify{\textbf{Final patch B}} \\
\midrule
\multicolumn{4}{l}{\modify{\textbf{Test outcomes (passed / total)}}} \\
\modify{\textbf{Deprecation warnings}} & \modify{0.375} & \gain{$\checkmark$~(3/3)} & \loss{$\times$~(0/3)} \\
\modify{\textbf{Module namespace}} & \modify{0.375} & \loss{$\times$~(0/1)} & \gain{$\checkmark$~(1/1)} \\
\modify{\textbf{Block construction}} & \modify{0.250} & \loss{$\times$~(0/1)} & \gain{$\checkmark$~(1/1)} \\
\modify{\textbf{Other shared tests}} & \modify{0.000} & \modify{4/4} & \modify{4/4} \\
\multicolumn{2}{l}{\modify{\textbf{Total tests passed}}} & \modify{7/9} & \modify{6/9} \\
\midrule
\multicolumn{4}{l}{\modify{\textbf{Final reward}}} \\
\multicolumn{2}{l}{\modify{\textbf{PassRate}}} & \modify{\textbf{0.778}} & \modify{0.667} \\
\multicolumn{2}{l}{\modify{\textbf{CodeLift}}} & \modify{0.375} & \modify{\textbf{0.625}} \\
\bottomrule
\end{tabular}
\end{table}

\modify{Table~\ref{tab:case-balancing} shows that raw PassRate score and CodeLift's behavior balanced score rank
the two final patches differently. PassRate favors A because its three
passing warning tests outnumber B's two passing tests for namespace and
block construction. CodeLift groups the tests and constructs three separate groups for warning, namespace, and block construction, with the weights of 0.375, 0.375, and 0.250. The other four
tests pass on every final patch and receive zero weight. Thus, CodeLift gives B a higher final reward for satisfying two
of the three behaviors, even though A passes more tests.
We examine how groups and weights change during training in
Appendix~\ref{app:curriculum-details}.}

\noindent\begin{minipage}{\linewidth}
\subsection{\modify{Snapshot slicing for intermediate credit}}
\label{app:case-credit}
\modify{In this subsection, we discuss two issue cases. The first asks
NumPy's \texttt{NumpyVersion} to parse versions such as \texttt{10.0.1};
the second asks aiohttp to reject chunked encoding for HTTP/1.0. We use
their recorded rollouts to compare how final-only credit and snapshot
slicing assign credit to intermediate and final patches
(Table~\ref{tab:case-slicing}). For each comparison, the state scores and
peer baseline are the same.}
\end{minipage}

\begin{table}[!htb]
\centering
\caption{\modify{\textbf{Intermediate and final credit in two recorded rollouts.}}}
\label{tab:case-slicing}
\begin{tabular}{llccc}
\toprule
\modify{\textbf{Case}} & \modify{\textbf{State}} & \modify{\shortstack{\textbf{Tests passed}}} & \modify{\shortstack{\textbf{Score} $S(s_i)$}} & \modify{\shortstack{\textbf{Slice credit $A_i^j$}}} \\
\midrule
\modify{\textbf{NumPy}} & \modify{\textbf{Intermediate snapshot}} & \modify{8/8} & \modify{1.000} & \gain{+0.143} \\
 & \modify{\textbf{Final patch}} & \modify{3/8} & \modify{0.333} & \loss{-0.524} \\
\midrule
\modify{\textbf{aiohttp}} & \modify{\textbf{Intermediate snapshot}} & \modify{63/66} & \modify{0.000} & \loss{-0.857} \\
 & \modify{\textbf{Final patch}} & \modify{66/66} & \modify{1.000} & \gain{+0.143} \\
\bottomrule
\end{tabular}
\end{table}

\paragraph{\modify{NumPy: a repair followed by regression.}}
\modify{The agent first modifies the regular expression to allow multiple
digits in the major version. This intermediate patch passes all eight
tests. However, the agent later rewrites the full pattern, breaking five
prerelease and development tests. With a final score of $1/3$ and a peer
baseline of $6/7$, final-only credit assigns $-11/21$ to both slices,
including the earlier successful repair. With snapshot slicing, CodeLift
gives the earlier repair $+1/7$ and keeps $-11/21$ for the final rewrite.}

\paragraph{\modify{aiohttp: a regression followed by repair.}}
\modify{The original code passes 65 of 66 tests. To reject HTTP/1.0 chunked
encoding, the agent adds a version check using \texttt{HttpVersion11} but
omits its import. This causes a \texttt{NameError}, leaving 63 of 66 tests
passing. The agent then adds the import while keeping the version check,
and the final patch passes all 66 tests.

With a peer baseline of $6/7$, final-only credit assigns $+1/7$ to both
slices, including the one that introduced the error. Although the
intermediate patch passes 63 tests, it fails every test group with
positive weight, giving it a score of zero. CodeLift therefore assigns
$-6/7$ to this intermediate slice and $+1/7$ to the final slice. The
negative credit applies to all actions in the intermediate slice,
including the version check retained in the final repair.}

\section{Prompt Templates}
\label{app:prompts}
We use the same system and user prompt templates across methods
(Figures~\ref{fig:agent-prompt}--\ref{fig:agent-prompt-cont}). The user prompt
inserts each issue at \texttt{\{problem\_statement\}}. For Pro, this includes
its requirements and interface specifications. We show the template with
\texttt{/testbed}, replacing it with \texttt{/app} for Pro.
Tool definitions are supplied separately by the agent interface.

\begingroup
\tcbset{agentprompt/.style={
  colback=gray!5,
  colframe=gray!70!black,
  coltitle=white,
  fonttitle=\normalfont,
  listing only,
  listing options={
    basicstyle=\ttfamily\color{black},
    columns=fullflexible,
    keepspaces=true,
    breaklines=true,
    breakatwhitespace=true,
    breakautoindent=false,
    breakindent=0pt,
    showstringspaces=false,
    upquote=true
  }
}}
\begin{figure}[!p]
\centering
\begin{tcblisting}{agentprompt,title=System prompt}
You are a helpful assistant that can interact with a computer to solve tasks.
\end{tcblisting}
\begin{tcblisting}{agentprompt,title=User prompt (1/3)}
<uploaded_files>
/testbed
</uploaded_files>
I have uploaded a python code repository in the /testbed directory. You can explore and modify files using the available tools. Consider the following issue description:

<issue_description>
{problem_statement}
</issue_description>

Can you help me implement the necessary changes to the repository to fix the <issue_description>?
I have already taken care of all changes to any of the test files described in the <issue_description>. This means you DON'T have to modify the testing logic or any of the tests in any way!
Also the development Python environment is already set up for you (i.e., all dependencies already installed), so you don't need to install other packages.
Your task is to make the minimal changes to non-test files in the /testbed directory to ensure the <issue_description> is satisfied.

Follow these steps to resolve the issue:
1. First, explore the codebase to locate and understand the code relevant to the <issue_description>. 
- Use efficient search commands to identify key files and functions.  
- You should err on the side of caution and look at various relevant files and build your understanding of 
    - how the code works
    - what are the expected behaviors and edge cases
    - what are the potential root causes for the given issue

2. Assess whether you can reproduce the issue:
- Create a script at '/testbed/reproduce_issue.py' that demonstrates the error.
- Execute this script to confirm the error behavior.
- You should reproduce the issue before fixing it.
- Your reproduction script should also assert the expected behavior for the fixed code. 

\end{tcblisting}
\caption{{System and user prompt templates (part 1 of 3).}}
\label{fig:agent-prompt}
\end{figure}
\begin{figure}[!p]
\centering
\begin{tcblisting}{agentprompt,title=User prompt (2/3)}
3. Analyze the root cause:
- Identify the underlying problem based on your code exploration and reproduction results.
- Critically analyze different potential approaches to fix the issue. 
- You NEED to explicitly reason about multiple approaches to fix the issue. Next, find the most elegant and effective solution among them considering the tradeoffs (correctness, generality, side effects, etc.).
- You would need to reason about execution paths, edge cases, and other potential issues. You should look at the unit tests to understand the expected behavior of the relevant code.

4. Implement your solution:
- Make targeted changes to the necessary files following idiomatic code patterns once you determine the root cause.
- You should be thorough and methodical.

5. Verify your solution:
- Rerun your reproduction script to confirm the error is fixed.
- If verification fails, iterate on your solution until successful. If you identify the reproduction script is buggy, adjust it as needed.

6. Run unit tests:
- Find and run the relevant unit tests relevant to the performed fix.
- You should run the unit tests to ensure your solution is correct and does not cause any regressions.
- In cases where the unit tests are do not pass, you should consider whether the unit tests does not reflect the *new* expected behavior of the code. If so, you can test it by writing additional edge test cases.
- Use the existing test runner to run the unit tests you identify as relevant to the changes you made. For example:
    - `python -m pytest -xvs sympy/physics/units/tests/test_dimensions_transcendental.py`
    - `python -m pytest tests/test_domain_py.py::test_pymethod_options`
    - `./tests/runtests.py constraints.tests.CheckConstraintTests -v 2`
- RUN ALL relevant unit tests to ensure your solution is correct and does not cause any regressions.
- DO NOT MODIFY any of the existing unit tests. You can add new edge test cases in a separate file if needed BUT DO NOT MODIFY THE EXISTING TESTS.

\end{tcblisting}
\caption{User prompt template (part 2 of 3).}
\label{fig:agent-prompt-middle}
\end{figure}
\begin{figure}[!p]
\centering
\begin{tcblisting}{agentprompt,title=User prompt (3/3)}
7. Test edge cases:
- Identify potential edge cases that might challenge your solution.
- Create additional test cases in a separate file '/testbed/edge_case_tests.py'.
- Execute these tests to verify your solution's robustness.
- You should run multiple rounds of edge cases. When creating edge cases:
    - Consider complex scenarios beyond the original issue description
    - Test for regressions to ensure existing functionality remains intact
    - At each round you should write multiple edge test cases in the same file to be efficient

8. Refine if necessary:
- If edge case testing reveals issues, refine your solution accordingly.
- Ensure your final implementation handles all identified scenarios correctly.
- Document any assumptions or limitations of your solution.

9. Submit your solution:
- Once you have verified your solution, submit your solution using the `submit` tool.

A successful resolution means:
- The specific error/issue described no longer occurs
- Your changes maintain compatibility with existing functionality
- Edge cases are properly handled
\end{tcblisting}
\caption{User prompt template (part 3 of 3).}
\label{fig:agent-prompt-cont}
\end{figure}
\endgroup

\end{document}